%% file: paper.tex
\documentclass[runningheads]{llncs}

\usepackage[T1]{fontenc}

\usepackage{graphicx}
\usepackage{multirow}
\usepackage{booktabs}
\usepackage{amsmath}
\usepackage{siunitx}
\usepackage{bbding}

\begin{document}

\title{Few-Shot Biomedical Relation Extraction with Large Language Models: A Viable Alternative to Supervised Learning?}
\titlerunning{Few-Shot BioRE with LLMs: A Viable Alternative to Supervised Learning?}

\author{
Jakob Mraz\Envelope\inst{1}\orcidID{0000-0001-7157-5264} \and 
Tomaž Curk\inst{1}\orcidID{0000-0003-4888-7256} \and 
Blaž~Zupan\inst{1,2}\orcidID{0000-0002-5864-7056}
}

\authorrunning{J. Mraz et al.}

\institute{
University of Ljubljana, Ljubljana, Slovenia\\
Baylor College of Medicine, Houston, Texas, USA
}

\maketitle

\begin{abstract}

Biomedical relation extraction (BioRE) is a key step in transforming biomedical literature into structured knowledge. Most existing approaches rely on supervised models trained on costly annotated datasets, limiting their scalability and adaptability across relation types and domains. We investigate few-shot BioRE using prompt-based learning with large language models (LLMs) and compare two task formulations: pairwise classification, which predicts relations for individual entity pairs, and joint generation, which extracts multiple relations in a single model call. Experiments on the BioREDirect dataset reveal a clear precision–recall trade-off. Pairwise classification achieves higher recall, whereas joint generation is more precise and computationally efficient. The best-performing model achieves a micro-F1 score of 0.44, substantially outperforming previous few-shot results (0.34) while remaining below the supervised baseline (0.56). Much of this gap is attributable to a single ambiguously defined relation type. When evaluated using macro-F1, which better captures performance across relation types in an imbalanced setting, prompt-based approaches outperform the supervised baseline (0.45 vs. 0.38), particularly on rare relation types. These findings highlight the potential of LLMs for BioRE in low-resource settings and underscore the importance of well-defined relation schemas.

\end{abstract}

\section{Introduction}

The volume of scientific publications has expanded rapidly in recent decades, leading to an unprecedented accumulation of written knowledge. This trend is particularly pronounced in the biomedical field, where the PubMed database alone indexes more than 1.5 million new articles annually~\cite{novoa_pmidigest_2023}, not including other sources of biomedical text such as clinical notes and electronic health records. While this growth creates new opportunities for discovery, it also renders manual analysis and knowledge integration increasingly infeasible.

\begin{figure}[t]
    \centering
    \includegraphics[width=\linewidth]{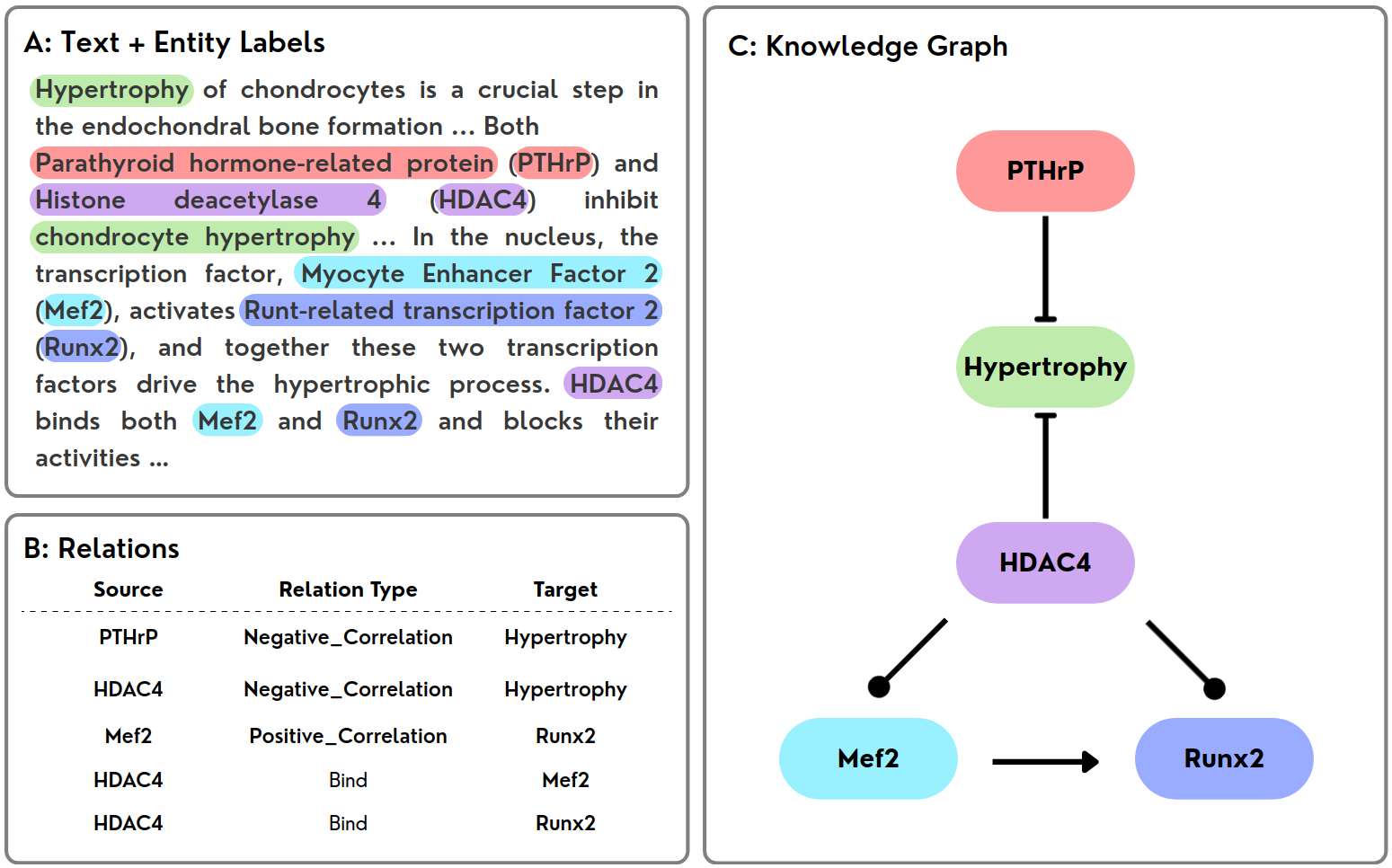}
    \caption{
    Conversion of unstructured text into structured biomedical knowledge. Text is first processed using named entity recognition and normalization to identify biomedical entities and map them to standardized identifiers (A). Next, relation extraction is used to identify semantic relationships between these entities (B). The extracted entities and their relationships can be represented as a knowledge graph, where nodes correspond to entities and edges represent their relationships (C). The example shown in this figure is adapted from the BioREDirect~\cite{lai_enhancing_2025} dataset, which is used throughout our analyses.
    }
    \label{fig:ds_scheme}
\end{figure}

Natural language processing techniques address this challenge by converting unstructured text into structured representations. A knowledge graph (KG) is a widely adopted framework that encodes entities and their relationships in both human- and machine-readable form~\cite{ji_survey_2022}. Constructing KGs from text relies on three core information extraction tasks: named entity recognition (NER), which identifies entity mentions; named entity normalization (NEN), which maps mentions to standardized database identifiers; and relation extraction (RE), which determines semantic relationships between the extracted entities (Figure~\ref{fig:ds_scheme}). While modern NER and NEN approaches achieve strong performance with robust tools~\cite{sanger_hunflair2_2024}, RE remains challenging due to its sensitivity to complex language, long-range dependencies, and domain variability.

Biomedical RE (BioRE) methods typically rely on supervised learning with expert-annotated data, which is costly and labor-intensive to obtain, and often struggle to generalize to rare or novel relation types and to new domains~\cite{goyal_named_2025}. This has motivated interest in prompt-based methods with large language models (LLMs) that are better suited to low-resource settings and cross-domain generalization. Trained on vast text corpora, LLMs encode extensive linguistic and factual knowledge in their parameters, which can be leveraged during inference. The combination of learned knowledge and prompt-conditioned generation enables LLMs to perform a wide range of tasks without task-specific training~\cite{chen_benchmarking_2025}. Yet, reported few-shot performance of LLMs for BioRE varies considerably, ranging from substantial underperformance relative to fully supervised methods~\cite{jimenez_gutierrez_thinking_2022,liu_comprehensive_2025,zhang_study_2024} to competitive results~\cite{agrawal_large_2022,jahan_comprehensive_2024,zhao_zero-shot_2026}, which motivates further research.

In this work, we investigate how prompting strategies affect BioRE performance across two task formulations: \emph{pairwise classification}, in which the model predicts the relation between a single annotated entity pair at a time, and \emph{joint generation}, in which the model predicts multiple relations among all annotated entities in a single call. We compare these two paradigms in terms of extraction performance and computational efficiency; to our knowledge, such a systematic comparison has not yet been conducted for BioRE. We perform our analysis using the recent open-weight Gemma-4 and Qwen-3.5 model families.

\section{Related Work}
\label{sec:related_work}

RE is the task of identifying semantic relationships between entities in text. It encompasses several subtasks, including identifying relevant entity pairs, classifying their relation types, and determining relation directionality. In the biomedical domain, RE is essential for uncovering interactions among genes, proteins, diseases, and chemical compounds, supporting applications such as drug discovery, pathway analysis, and disease modeling.

\subsubsection{Rule-Based Methods.}

Early BioRE approaches relied on co-occurrence heuristics or manually designed rules based on lexical and syntactic patterns~\cite{ben_abacha_automatic_2011,huang_discovering_2004}. While some achieve high precision, these methods are typically labor-intensive and suffer from low recall and limited generalizability. In particular, they struggle to capture complex linguistic phenomena such as negation and long-range dependencies and often fail to transfer to new datasets or domains~\cite{diaz-garcia_survey_2025}.

\subsubsection{Supervised Learning.}

Modern BioRE approaches are predominantly based on supervised learning, where models are trained to predict relations from annotated examples. Early methods relied on traditional machine learning techniques such as support vector machines~\cite{peng_extracting_2018} and graph convolutional networks~\cite{zhao_biomedical_2021}, while more recent approaches are based on pre-trained language models (PLMs)~\cite{wei_relation_2020}.

PLM-based methods typically follow a two-stage paradigm. First, models are pre-trained on large unlabeled corpora using a language modeling objective. Second, they are fine-tuned on task-specific datasets using supervised objectives tailored to the extraction task. The most widely adopted architecture is Bidirectional Encoder Representations from Transformers (BERT)~\cite{devlin_bert_2019}, which has become the dominant backbone for modern BioRE systems. For example, Lai \textit{et al.}~\cite{lai_enhancing_2025} combined PubMedBERT with soft-prompt tuning and multi-task learning to achieve state-of-the-art results on the BioREDirect and BC5CDR datasets. To overcome BERT's 512-token input limit, the method employs chunking, allowing the model to leverage information from different sections of a document while simultaneously predicting relation type, directionality, and novelty.

Additionally, LLMs have been adapted to BioRE through fine-tuning. For example, Peng \textit{et al.}~\cite{peng_model_2024} investigated clinical relation extraction by comparing full model fine-tuning, soft-prompt tuning, and their combination. They showed consistent improvements across all settings, with the combined approach achieving the best performance. They further observed that the performance gap between fine-tuned and untuned models decreases with model size, suggesting that larger LLMs may perform BioRE effectively without task-specific training.

Despite strong performance, PLM- and LLM-based supervised approaches have several limitations. These models are computationally expensive to train and generalize poorly when trained on small or highly specialized datasets~\cite{zhao_comprehensive_2024}. Therefore, their effectiveness is heavily dependent on large, high-quality annotated datasets, which are costly and time-consuming to construct in the biomedical domain due to the need for expert annotation~\cite{shang_biomedical_2025}.

\subsubsection{Prompt-Based Learning.}

To address the limitations of supervised systems, prompt-based learning with LLMs, particularly zero- and few-shot approaches, has attracted increasing attention. As model architectures and training corpora continue to scale, LLMs demonstrate a growing ability to adapt to downstream tasks without task-specific fine-tuning~\cite{zhao_zero-shot_2026}. Prompts typically describe the extraction task, target relation types, and include a small number of demonstrations to guide the model towards the desired behavior. Recent work has also explored advanced prompting strategies such as chain-of-thought, question-answering, and self-verification to further improve performance~\cite{xu_large_2024}.

Existing work has primarily explored two BioRE task formulations: pairwise classification, which predicts relations for individual entity pairs, and joint generation, which extracts multiple relations in a single model call. Using pairwise classification, Zhao \textit{et al.}~\cite{zhao_zero-shot_2026} demonstrated that carefully designed prompts can achieve performance competitive with supervised methods. Their approach combines task instructions describing the target relations and extraction criteria with both positive and negative examples, yielding substantially higher recall than the corresponding supervised baselines, albeit at the cost of lower precision.

In contrast, Liu \textit{et al.}~\cite{liu_comprehensive_2025} evaluated several open-source LLMs using joint generation in both few-shot and fine-tuned settings. They found that larger models substantially outperform smaller ones and that parameter-efficient fine-tuning methods such as LoRA can partially close the performance gap. Nevertheless, even the strongest evaluated LLMs remained well below supervised BERT-based approaches. Notably, their prompts did not explicitly describe the target relation types, which may have limited the model's ability to distinguish them effectively.

Overall, prompt-based methods offer a flexible alternative to supervised systems, particularly in low-resource and cross-domain settings, as they can generalize from only a small number of examples. Their effectiveness depends strongly on prompt design, model architecture, and task formulation. Because LLM inference is substantially more costly than that of conventional supervised models, both extraction quality and computational efficiency must be considered when evaluating BioRE systems. This motivates a systematic evaluation of prompt-based methods and LLM architectures for BioRE.

\section{Evaluation Setup}
\label{sec:setup}

We evaluated prompt-based learning for BioRE in a few-shot setting using recent LLMs, comparing pairwise classification and joint generation task formulations. Our work considers both extraction quality and computational efficiency and relates the results to existing supervised and prompt-based baselines.

Unlike most prior work that focuses on sentence-level extraction~\cite{zhao_comprehensive_2024}, we investigate document-level BioRE. This setting is more challenging due to the larger number of entity pairs and long-range dependencies, but better reflects real-world texts, where relations often span sentence boundaries~\cite{yao_docred_2019}.

\subsection{Dataset}

\input{tables/ds_types}

Experiments were conducted on the BioREDirect dataset~\cite{lai_enhancing_2025}, which consists of PubMed abstracts hand-annotated with six entity types and eight relation types, as described in Table~\ref{tab:ds_types}. Relation annotations are not provided for Species and CellLine entities or for Gene–Variant and Disease–Disease pairs.
The relation type distribution is highly imbalanced, with \textit{Association}, \textit{Positive Correlation}, and \textit{Negative Correlation} accounting for more than 95\% of all relation instances. Our evaluations were performed on the test split, which contains 400 abstracts and 6,036 annotated relation instances.

\subsection{Models}
We employed models from two recent open-source LLM families, Gemma-4\footnote{\url{https://huggingface.co/collections/google/gemma-4}} and Qwen-3.5\footnote{\url{https://huggingface.co/collections/Qwen/qwen35}}, which have demonstrated strong performance across a range of benchmarks. We focused on medium-sized variants (25--35B parameters) that can be reliably deployed on a single NVIDIA DGX Spark system with 128 GB of unified memory.

The evaluated models include both dense and mixture-of-experts (MoE) architectures. Dense models utilize all parameters for every input token, leading to consistent but computationally expensive inference. In contrast, MoE models activate only a subset of parameters (experts) per token, reducing computational costs while maintaining high total parameter counts. 

All models were evaluated in full precision (BF16) using deterministic decoding with temperature, presence penalty, and frequency penalty set to 0, top-p and top-k set to 1, and a fixed random seed. Experiments were conducted both with and without \emph{reasoning} enabled, which allows the model to generate intermediate reasoning steps before producing a final response, potentially improving performance on complex tasks at the cost of increased inference latency.

\subsection{Tasks}

\begin{figure}[t]
    \centering
    \includegraphics[width=\linewidth]{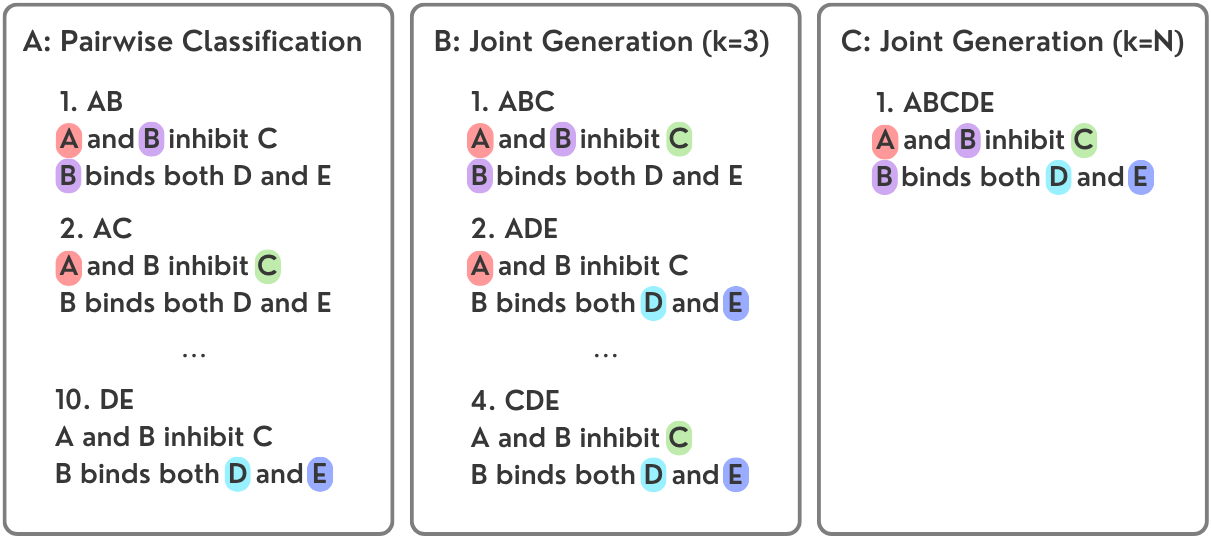}
    \caption{
    Comparison of BioRE task formulations for an input text with $N=5$ entities. The color highlights represent the annotated entities. In the pairwise classification setting, each entity pair is evaluated independently, resulting in 10 model calls. In the joint generation setting, each call contains a subset of $k$ entities, constructed such that every entity pair co-occurs in at least one call. With $k=3$, all pairs are covered in 4 calls, while $k=N$ reduces the task to a single call.
    }
    \label{fig:task}
\end{figure}

We performed BioRE using prompt-based learning. Each prompt contained task instructions, relation type descriptions, and a small set of few-shot examples. Relation type descriptions were derived from the original annotation guidelines used during dataset construction. The model was instructed to extract only relations explicitly stated in the text; co-mentions, indirect reasoning, and ambiguous relation directionality were treated as negative instances. Few-shot examples were sampled from the train split and covered all relation types. The input text, annotated with the entities of interest, was appended to the end of the prompt.

We evaluated two task formulations (Figure~\ref{fig:task}):

\begin{itemize}
    \item \textbf{Pairwise Classification (Cls)}: The model receives the input text annotated with a single entity pair and predicts the relation type between them, or \emph{None} if no relation is supported. This procedure is repeated independently for each candidate entity pair.

    \item \textbf{Joint Generation (Gen)}: The model receives the input text annotated with multiple entities and generates all supported $(\text{head}, \text{tail}, \text{relation})$ triples in a single call. If no valid relations are present, the model outputs \emph{None}.
\end{itemize}

The BioREDirect annotation schema was enforced throughout evaluation. Pairwise classification considered only valid entity type combinations, and generated predictions violating entity-pair or relation-type constraints were discarded.

To study the trade-off between extraction quality and computational efficiency, we additionally varied the context scope in the joint generation setting using a parameter $k$, which defines the number of entities included in a single model call. At $k=N$, all entities in an abstract are processed together in a single call. For smaller values of $k$, entities are divided into overlapping subsets constructed so that every entity pair appears in at least one subset. We used a greedy approximation algorithm to construct a near-minimal set of subsets covering all entity pairs. Predictions from overlapping subsets were aggregated by majority vote for each entity pair, with ties discarded as ambiguous.

\subsection{Metrics}

Extraction performance was evaluated using precision, recall, and F1 scores based on exact matching against gold annotations. Metrics were computed for both entity pair (EP) and relation type (RT) extraction. Due to substantial class imbalance, we report both micro- and macro-F1 scores.

In addition, we evaluated computational efficiency using weighted token cost (Cost). Let $T_{in}$ and $T_{out}$ denote the number of input and output tokens required per extraction run, and let $c_i$ and $c_o$ denote their relative processing costs. Total cost is computed as

\begin{equation}
\textit{Cost} = c_i T_{in} + c_o T_{out}
\end{equation}

The relative costs $c_i$ and $c_o$ were estimated empirically from model throughput benchmarks. Input and output throughput were measured separately as the average number of tokens processed per second during prompt prefill and token decoding, respectively, using repeated sequential inference with synthetic prompts containing unique prefixes to prevent prompt-cache reuse. Costs were defined as the inverse throughput, assigning higher weights to slower operations.

\section{Results}
\label{sec:results}

\input{tables/perform}

Table~\ref{tab:perform} summarizes extraction performance across model architectures and task formulations. The highest EP+RT micro-F1 score of 0.44 is achieved by Qwen3.5-27B in the classification setting. Our approaches substantially outperform the few-shot results reported by Liu \textit{et al.}~\cite{liu_comprehensive_2025} but remain below the supervised baselines of Lai \textit{et al.}~\cite{lai_enhancing_2025}. This is largely attributable to performance on the \textit{Association} class. Considering macro-F1 scores, the performance gap narrows considerably: the best result is achieved by Gemma-4 models in the generation setting, with an EP+RT macro-F1 score of 0.45, compared with 0.38 for the BERT baseline.

The \textit{Association} class is inherently ambiguous, serving as a fallback label for cases where a more specific relation type cannot be determined confidently during annotation. Rather than predicting this generic label, our LLM-based approach frequently assigned more specific mechanistic relation types, such as \textit{Positive}, \textit{Negative}, or \textit{Bind}. This behavior suggests a tendency to infer regulatory or physical interactions from statements describing only general associations, thereby overinterpreting the available evidence. Consequently, the ambiguity of the \textit{Association} class appears to pose a greater challenge for instruction-driven LLMs than for supervised models trained directly on the annotated corpus.

\begin{figure}[t]
    \centering
    \includegraphics[width=\linewidth]{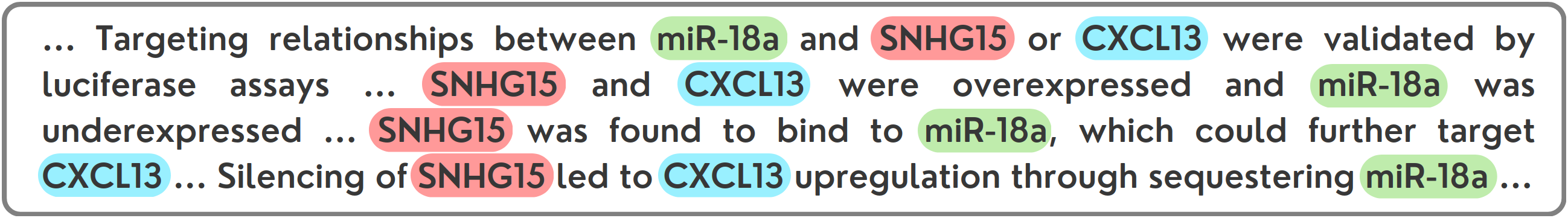}
    \caption{
    Example of BioREDirect annotation ambiguity: the abstract contains three entities – SNHG15 (lncRNA), miR-18a (miRNA), and CXCL13 (protein). All pairwise relations are labeled as \textit{Association} while the text provides evidence for more specific relations. The relation between SNHG15 and miR-18a is explicitly described as \textit{Bind} (“SNHG15 was found to bind to miR-18a”), while the statement “Silencing of SNHG15 led to CXCL13 upregulation” suggests a negative regulatory relation between SNHG15 and CXCL13. This illustrates how the \textit{Association} labels in the gold dataset can mask more specific relationships described in the text.
    }
    \label{fig:ds_mistake}
\end{figure}

A closer inspection of the misclassified instances, however, suggests that some \textit{Association} annotations could reasonably be assigned more specific relation types. Figure~\ref{fig:ds_mistake} illustrates one such case where our model predicts more specific relation types and is penalized when evaluated against the gold annotation. While this does not necessarily indicate an annotation error, it highlights the subjective nature of relation categorization in borderline cases. As a result, part of the observed performance difference may reflect limitations of the annotation scheme rather than extraction capability, suggesting that the true performance gap may be smaller than the BioREDirect scores indicate.

\input{tables/perform_per_class}

Table~\ref{tab:perform_per_class} shows extraction performance per relation type for gemma-4-31B-it. The model outperforms the supervised baseline on several low-frequency relation types, including \textit{Bind}, \textit{Comparison}, and \textit{Cotreatment}. This suggests that LLMs are less affected by class imbalance when relation definitions are sufficiently specific and semantically distinct, representing a key advantage of LLM-based approaches for BioRE. These findings should nevertheless be interpreted with caution, as the number of evaluation instances for these relation types, particularly \textit{Comparison} and \textit{Cotreatment}, is small, and the corresponding performance estimates may therefore be subject to substantial variance.

\subsubsection{Task Formulations.}

\begin{figure}[t]
    \centering
    \includegraphics[width=0.95\linewidth]{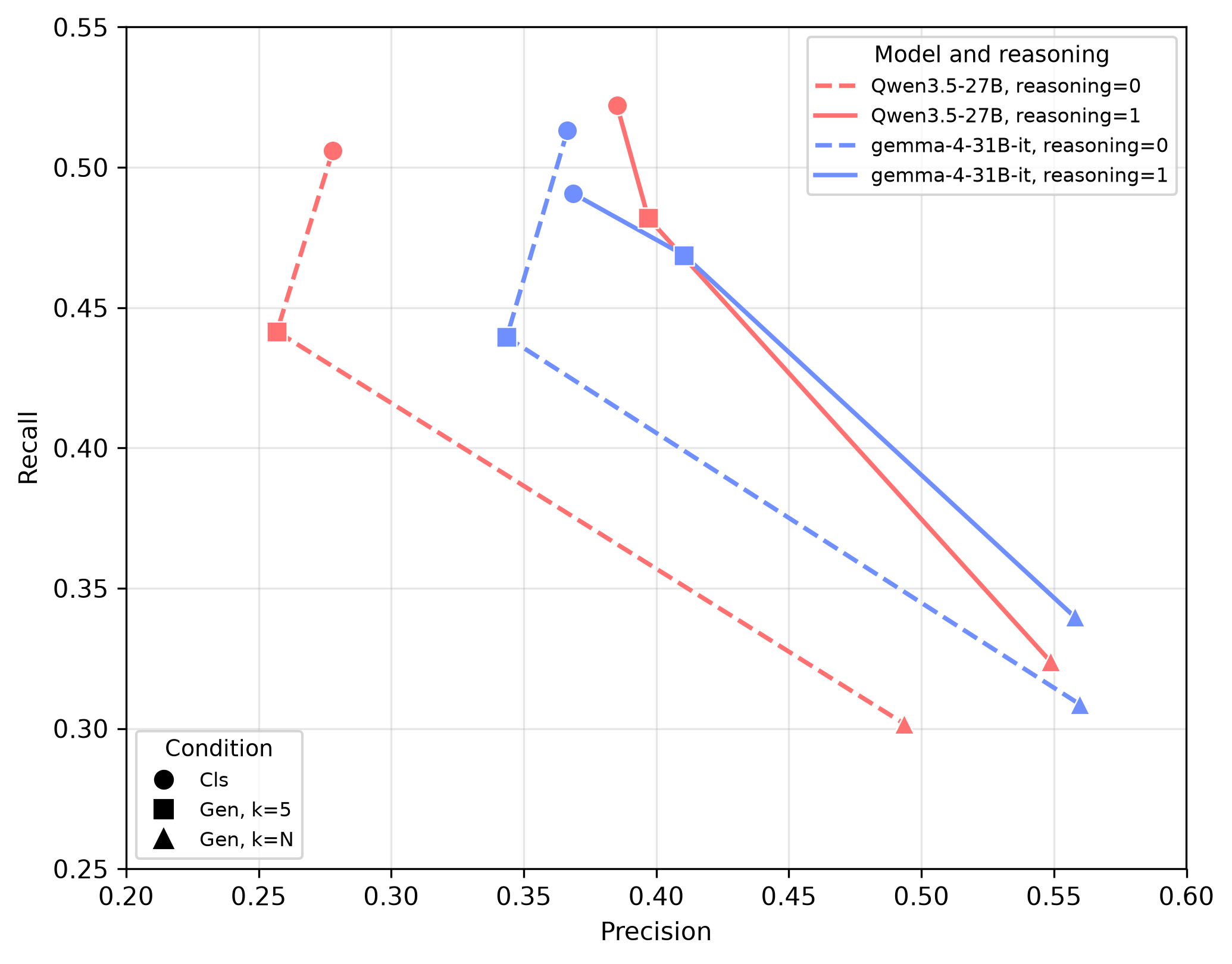}
    \caption{
    Precision--recall trade-off for the EP+RT task. Pairwise classification (circles) consistently achieves higher recall, whereas joint generation (squares: k=5; triangles: k=N) tends to achieve higher precision at the expense of recall. The trends are shown for the two best-performing models, gemma-4-31B-it and Qwen3.5-27B, with colors denoting models and line styles indicating reasoning settings.
    }
    \label{fig:tradeoff}
\end{figure}

The two task formulations exhibit a clear precision--recall trade-off. As shown in Figure~\ref{fig:tradeoff}, pairwise classification consistently achieves higher recall, whereas joint generation yields higher precision at the expense of recall. The same trend is observed across all evaluated models and is consistent with the findings of Zhao \textit{et al.}~\cite{zhao_zero-shot_2026}.

These differences stem from distinct prediction behaviors. Pairwise classification achieves high coverage but also produces more false positives, often predicting direct relations between entities that are connected only through intermediate nodes in the gold relation graph. This effectively collapses multi-step biological pathways into direct pairwise relations, particularly in pathway descriptions where entities co-occur within the same mechanistic context. Joint generation, in contrast, adopts a more conservative strategy, typically extracting only the most salient relations. While this substantially reduces false positives, it also increases false negatives, especially in abstracts with many annotated relations. Overall, pairwise classification tends to overinterpret context, whereas joint generation is primarily limited by incomplete relation coverage.

Introducing $k$-constrained generation provides a mechanism for navigating this precision--recall trade-off. By limiting the number of entity pairs considered in a single generation step ($k<N$), the model achieves precision levels that remain higher than for pairwise classification while recovering some of the recall lost in unconstrained generation. As a result, $k$ acts as a controllable parameter that allows practitioners to balance extraction conservativeness against relation coverage according to task requirements.

\input{tables/usage}
The relative costs presented in Table~\ref{tab:usage} reveal a substantial efficiency advantage for joint generation, reducing computational costs by up to 25$\times$ compared to pairwise classification. This difference arises because classification requires a separate model call for each entity pair, whereas generation can extract multiple relations at once. The $k$-constrained generation provides an intermediate cost profile. By extracting relations for multiple entity pairs per call, it remains substantially more efficient than pairwise classification, while the need for multiple calls to cover all pairs results in higher costs than full joint generation.

\subsubsection{Model architectures and reasoning.}
No single model family consistently outperforms the others across all settings. In the pairwise classification setting, Qwen-3.5 models achieve the strongest results in both micro- and macro-F1, whereas Gemma-4 models perform best in the joint generation setting.

Enabling reasoning generally improves extraction performance, likely by facilitating more effective processing of complex relations. The gains are particularly pronounced for MoE models, which appear to benefit more from reasoning than dense architectures. These improvements come at a substantial computational cost. Reasoning significantly increases the number of generated output tokens, which are considerably more expensive than input tokens (see the $c$ values in Table~\ref{tab:usage}), thereby increasing inference costs considerably.

Computational efficiency also varies across architectures. MoE models are consistently more cost-efficient than dense models despite producing a similar number of output tokens. By activating only a subset of parameters for each token, MoE architectures reduce inference costs substantially. Nevertheless, dense models retain a slight performance advantage, particularly when reasoning is disabled.

\subsubsection{Limitations and future work.}
Several avenues may further improve extraction performance. Larger and biomedical domain-specific LLMs are promising candidates, while uncertainty-aware prediction strategies that leverage token probabilities could help filter low-confidence predictions in the classification setting. Additional gains may come from advanced prompting techniques, such as chain-of-thought reasoning, and task-specific preprocessing methods, including entity masking, synonym replacement, and irrelevant entity filtering. From an efficiency perspective, inference costs could be reduced through model quantization, improving deployment practicality without requiring architectural changes.

A limitation of this study is that complete reproducibility cannot be guaranteed. Although all experiments used deterministic decoding with a temperature of zero, small variations in model outputs were observed across runs. This behavior is a known characteristic of LLM inference and likely arises from low-level computational differences during execution. Consequently, evaluation scores may exhibit minor fluctuations. Repeated evaluations and confidence intervals should therefore be used to better characterize performance variability.

Finally, our evaluation focused exclusively on EP and RT extraction and did not assess relation directionality or novelty, which are also annotated in the dataset. Incorporating these tasks would provide a more comprehensive evaluation of BioRE performance. Future studies could additionally compare within- and cross-sentence relations to further disentangle the strengths of few-shot and supervised approaches. Given their ability to process longer contexts, LLMs may outperform BERT-based models on cross-sentence relation extraction.

\section{Conclusion}

In this work, we evaluated modern medium-sized LLMs for few-shot biomedical relation extraction. Our results demonstrate that they provide a viable alternative to supervised systems, particularly in settings with limited annotated data and well-defined relation semantics. Although the proposed approaches remain below the supervised baseline in terms of micro-F1, this gap appears to be strongly influenced by dataset imbalance and ambiguity in the \textit{Association} class. When considering macro-F1, which better reflects performance across relation types, LLM-based approaches outperform the supervised baseline, particularly on low-frequency relations.

Across task formulations, classification and generation achieved broadly comparable F1 scores, with performance differences primarily reflecting a precision--recall trade-off rather than a clear superiority of either approach. Pairwise classification consistently achieved higher recall and is therefore better suited to applications requiring maximal relation coverage. In contrast, joint generation yielded higher precision while incurring substantially lower computational costs, making it attractive for scenarios where extraction quality and efficiency are prioritized. The proposed k-constrained generation setting provides an effective intermediate alternative, enabling practitioners to balance precision, recall, and inference cost according to downstream task requirements.

\section*{Data and Code Availability}

The source code and input data used in this study are available at \url{https://github.com/jkbmrz/few-shot-biore}.

\section*{Funding}

This work was supported by core funding \mbox{P2-0209} (Artificial Intelligence and Intelligent Systems) and project grants \mbox{GC-0001} (Artificial Intelligence for Science) and \mbox{L2-60154} (Explainable Foundation Models for Human Gene Expression), all funded by Slovenian Research and Innovation Agency.

\bibliographystyle{abbrv} % abbreviated references
\bibliography{references}  % file name without .bib

\end{document}

%% file: tables/ds_types.tex
\begin{table}[t]
\centering
\small
\caption{\textbf{BioREDirect entity and relation type labels}. Test split counts are reported. Species and CellLine entity types are omitted because their relations are excluded from the annotation schema.
%Moreover, Gene--Variant and Disease--Disease relations are excluded from the annotation schema.
}
\label{tab:ds_types}

\renewcommand{\arraystretch}{1.1}
\setlength{\tabcolsep}{4pt}

\begin{tabular}{p{0.34\linewidth}rp{0.50\linewidth}}
\hline
\textbf{Type} & \textbf{Counts} & \textbf{Description} \\
\hline

\multicolumn{3}{l}{\textit{Entity Types}} \\
GeneOrGeneProduct & 5,728 & Genes/proteins (e.g., \textit{TP53}, EGFR) \\
DiseaseOrPhenotypicFeature & 3,641 & Diseases/phenotypes (e.g., breast cancer) \\
ChemicalEntity & 2,592 & Drugs/chemicals (e.g., aspirin, glucose) \\
SequenceVariant & 1,774 & Genetic variants (e.g., \textit{BRAF} V600E) \\

\hline
\multicolumn{3}{l}{\textit{Relation Types}} \\
Association & 2,759 & Relation without clear mechanism \\
Positive Correlation & 1,751 & Promotes, induces, or increases \\
Negative Correlation & 1,192 & Inhibits, treats, or decreases \\
Cotreatment & 172 & Combination drug therapy \\
Bind & 136 & Direct molecular binding \\
Comparison & 13 & Explicit comparison only \\
Conversion & 13 & One chemical converts to another \\
Drug Interaction & 0 & Pharmacological drug interaction \\
\hline

\end{tabular}
\end{table}

%% file: tables/perform.tex
\begin{table*}[t]
\centering
\small

\caption{\textbf{General BioRE performance}. F1 scores for EP and EP+RT tasks with reasoning disabled/enabled. For EP+RT condition, both Micro-F1 and Macro-F1 are reported due to relation type imbalance. Best results are shown in bold.}

\label{tab:perform}

\renewcommand{\arraystretch}{1.15}
\setlength{\tabcolsep}{12pt}

\begin{tabular*}{\textwidth}{@{\extracolsep{\fill}} l c c c @{}}
\hline
\textbf{Model}
& \textbf{EP}
& \multicolumn{2}{c}{\textbf{EP+RT}} \\
\cline{3-4}
& 
&
\textbf{Micro-F1}
&
\textbf{Macro-F1} \\
\hline

\multicolumn{4}{l}{\textit{Pairwise Classification}} \\
\hline

gemma-4-31B-it
& \textbf{0.68} / 0.67
& 0.43 / 0.42
& 0.37 / 0.39 \\

gemma-4-26B-A4B-it
& 0.65 / 0.66
& 0.38 / 0.42
& 0.38 / 0.40 \\

Qwen3.5-27B
& 0.59 / 0.67
& 0.35 / \textbf{0.44}
& 0.31 / 0.41 \\

Qwen3.5-35B-A3B
& 0.17 / 0.67
& 0.11 / 0.43
& 0.16 / 0.40 \\

%\hline
%\multicolumn{4}{l}{\textit{Generation (k=5)}} \\
%\hline

%gemma-4-31B-it
%& TBA / TBA
%& TBA / TBA
%& TBA / TBA \\

%gemma-4-26B-A4B-it
%& TBA / TBA
%& TBA / TBA
%& TBA / TBA \\

%Qwen3.5-27B
%& TBA / TBA
%& TBA / TBA
%& TBA / TBA \\

%Qwen3.5-35B-A3B
%& TBA / TBA
%& TBA / TBA
%& TBA / TBA \\

\hline
\multicolumn{4}{l}{\textit{Joint Generation (k=N)}} \\
\hline

gemma-4-31B-it
& 0.60 / 0.63
& 0.40 / 0.42
& 0.39 / \textbf{0.45} \\

gemma-4-26B-A4B-it
& 0.53 / 0.57
& 0.34 / 0.39
& 0.36 / \textbf{0.45} \\

Qwen3.5-27B
& 0.59 / 0.61
& 0.37 / 0.41
& 0.37 / 0.43 \\

Qwen3.5-35B-A3B
& 0.56 / 0.62
& 0.34 / 0.42
& 0.34 / 0.42 \\

\hline
\multicolumn{4}{l}{\textit{Baselines}} \\
\hline

Deepseek-v3 + few-shot \cite{liu_comprehensive_2025}
& --
& 0.34
& -- \\

GPT-3.5 + fine-tune \cite{lai_enhancing_2025}
& 0.70
& 0.50
& -- \\

PubMedBERT \cite{lai_enhancing_2025}
& 0.75
& 0.56
& 0.38 \\

\hline
\end{tabular*}
\end{table*}

%% file: tables/perform_per_class.tex
\begin{table*}[t]
\centering
\small

\caption{\textbf{Relation-level BioRE performance}. F1 scores for EP+RT task. The number of instances for each relation type is shown for reference. Results are shown for Gemma-4-31B-it with reasoning enabled. Best results are shown in bold.}
\label{tab:perform_per_class}

\renewcommand{\arraystretch}{1.15}
\setlength{\tabcolsep}{4pt}

\begin{tabular*}{\textwidth}{
  @{\extracolsep{\fill}}
  l r c c c c
  @{}
}
\hline
\textbf{Relation Type}
& \textbf{Counts}
& \textbf{Cls}
& \multicolumn{2}{c}{\textbf{Gen}}
& \textbf{PubMedBERT} \\
\cline{4-5}
&
&
&
\textit{$k=5$}
&
\textit{$k=N$}
& \\
\hline

Association          & 2,759 & 0.27 & 0.29 & 0.20 & \textbf{0.54} \\
Positive Correlation & 1,751 & 0.44 & 0.46 & 0.48 & \textbf{0.56} \\
Negative Correlation & 1,192 & 0.55 & 0.57 & 0.61 & \textbf{0.66} \\
Cotreatment          &   172 & 0.73 & 0.72 & 0.72 & \textbf{0.74} \\
Bind                 &   136 & 0.50 & 0.53 & \textbf{0.59} & 0.47 \\
Comparison           &    13 & 0.13 & 0.17 & \textbf{0.34} & 0.25 \\
Conversion           &    13 & 0.53 & 0.48 & \textbf{0.69} & 0.00 \\

\hline
\end{tabular*}
\end{table*}

%% file: tables/usage.tex
\begin{table*}[t]
\centering
\small

\caption{\textbf{Cost analysis}. Relative input and output cost coefficients ($c_i$ and $c_o$) are estimated from the inverse throughput of each model. For easier comparison, all costs are normalized with respect to Gemma-4-31B-it (joint generation with $k=N$ and reasoning enabled). Cost values are reported both for reasoning disabled/enabled.}

\label{tab:usage}

\renewcommand{\arraystretch}{1.15}
\setlength{\tabcolsep}{3pt}

\begin{tabular*}{\textwidth}{
@{\extracolsep{\fill}}
l
c
c
c@{\hspace{-0.45em}}c@{\hspace{-0.45em}}c
c@{\hspace{-0.45em}}c@{\hspace{-0.45em}}c
c@{\hspace{-0.45em}}c@{\hspace{-0.45em}}c
@{}
}
\hline

\textbf{Model}
& \textbf{$c_i$}
& \textbf{$c_o$}
& \multicolumn{3}{c}{\textbf{Cls}}
& \multicolumn{6}{c}{\textbf{Gen}} \\
\cline{7-12}

&
&
&
\multicolumn{3}{c}{}
& \multicolumn{3}{c}{$k=5$}
& \multicolumn{3}{c}{$k=N$} \\

\hline

Gemma-4-31B-it
& 0.0010
& 0.27
& 0.12 & \makebox[0pt][c]{/} & 10.38
& 0.20 & \makebox[0pt][c]{/} & 5.58
& 0.04 & \makebox[0pt][c]{/} & 1.00 \\

Gemma-4-26B-A4B-it
& 0.0002
& 0.04
& 0.03 & \makebox[0pt][c]{/} & 2.34
& 0.03 & \makebox[0pt][c]{/} & 1.56
& 0.01 & \makebox[0pt][c]{/} & 0.21 \\

Qwen3.5-27B
& 0.0008
& 0.22
& 0.10 & \makebox[0pt][c]{/} & 33.83
& 0.91 & \makebox[0pt][c]{/} & 16.62
& 0.05 & \makebox[0pt][c]{/} & 1.72 \\

Qwen3.5-35B-A3B
& 0.0003
& 0.03
& 0.03 & \makebox[0pt][c]{/} & 5.95
& 0.06 & \makebox[0pt][c]{/} & 2.22
& 0.02 & \makebox[0pt][c]{/} & 0.23 \\

\hline
\end{tabular*}
\end{table*}